\pdfoutput=1

\documentclass[11pt]{article}

\usepackage{EMNLP2023}

\usepackage{times}
\usepackage{latexsym}
\usepackage{tcolorbox}

\usepackage[T1]{fontenc}

\usepackage[utf8]{inputenc}

\usepackage{microtype}

\usepackage{inconsolata}

\usepackage{amsmath}
\usepackage{amssymb}
\usepackage{mathtools}
\usepackage{amsthm}
\usepackage{multirow}
\usepackage{graphicx}
\usepackage{natbib}

\usepackage[textsize=tiny]{todonotes}
\usepackage{fancyhdr}
\usepackage{xcolor} 
\usepackage{algorithm}
\usepackage{algorithmic}

\newcommand{\selapp}{{\scshape MinSimilarity}}

\usepackage{hyperref}
\usepackage[utf8]{inputenc} 
\usepackage[T1]{fontenc}    
\usepackage{url}            
\usepackage{booktabs}       
\usepackage{amsfonts}       
\usepackage{nicefrac}       
\usepackage{microtype}      
\usepackage{xcolor}         

\usepackage[flushmargin,hang,para]{footmisc}
\author{
    Xiaobo Guo\thanks{These authors contributed equally to this work.}, 
    Jay Desai\footnotemark[1], 
    \textbf{Srinivasan H. Sengamedu} 
    \\
    \textbf{Amazon} 
    \\
    \texttt{\href{mailto:xiaobo.guo.gr@dartmouth.edu}{xiaobo.guo.gr@dartmouth.edu}} \\
    \texttt{\href{mailto:jdesa@amazon.com}{jdesa@amazon.com}} \\
    \texttt{\href{mailto:sengamed@amazon.com}{sengamed@amazon.com}} 
    \\
    \thanks{We want to thank Weijie Xu and Stephanie Eckman for discussion and edits to the paper.}
}

\begin{document}

\title{JADS: A Framework for Self-supervised \\ Joint Aspect Discovery and Summarization}
\maketitle

\begin{abstract}

To generate summaries that include multiple aspects or topics for text documents, most approaches use clustering or topic modeling to group relevant sentences and then generate a summary for each group. These approaches struggle to optimize the summarization and clustering algorithms jointly. On the other hand, aspect-based summarization requires known aspects. Our solution integrates topic discovery and summarization into a single step. Given text data, our Joint Aspect Discovery and Summarization algorithm (JADS) discovers aspects from the input and generates a summary of the topics, in one step. We propose a self-supervised framework that creates a labeled dataset by first mixing sentences from multiple documents (e.g., CNN/DailyMail articles) as the input and then uses the article summaries from the mixture as the labels. The JADS model outperforms the two-step baselines. With pretraining, the model achieves better performance and stability. Furthermore, embeddings derived from JADS exhibit superior clustering capabilities. Our proposed method achieves higher semantic alignment with ground truth and is factual.

\end{abstract}

\begin{figure*}
\centering
\includegraphics[scale=0.53]{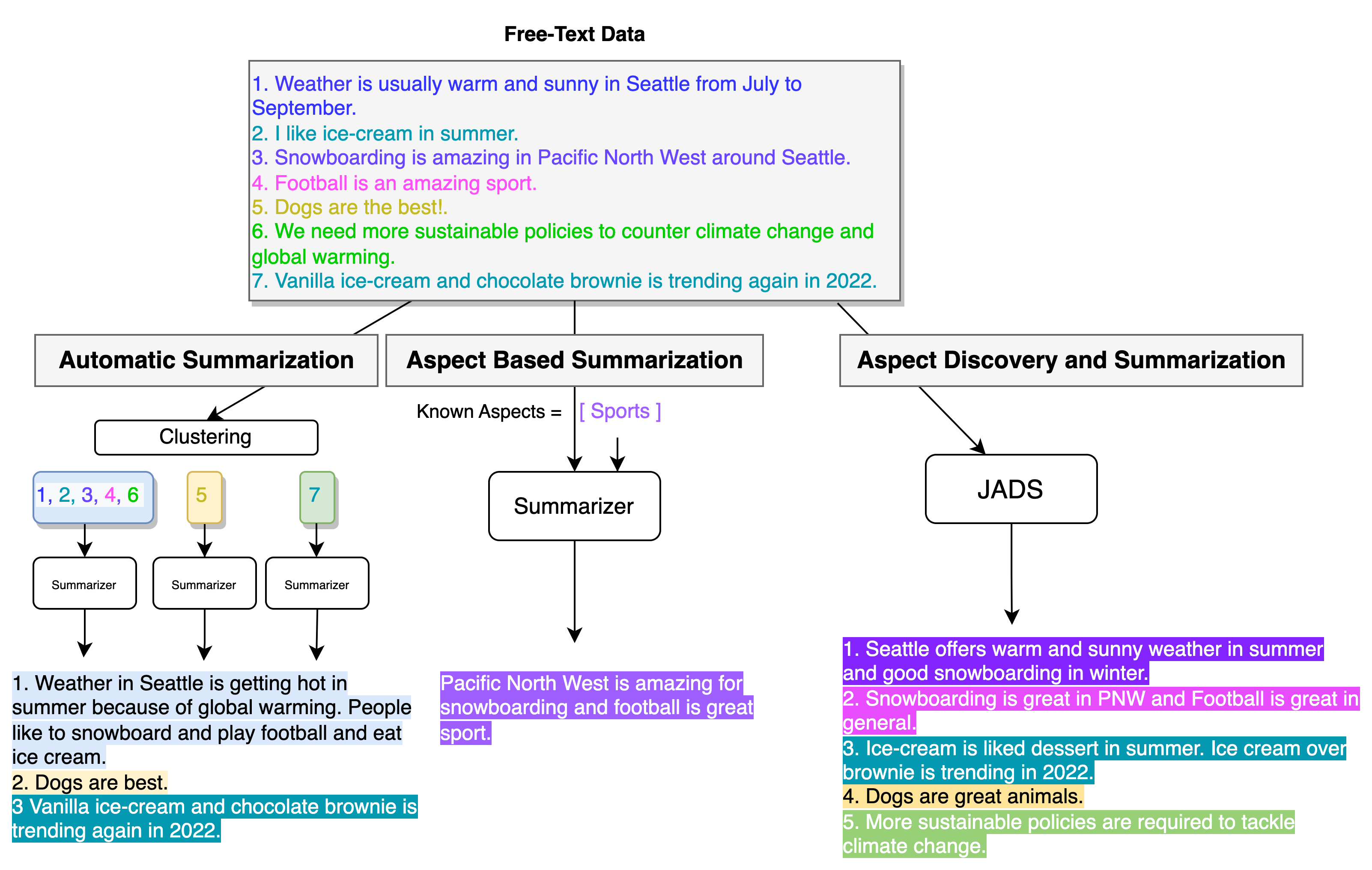}
\caption{JADS task compared to other summarization tasks.}
\label{fig:JADS_task}
\end{figure*}

\section{Introduction}

    Text summarization aims to generate condensed text representations that capture the salient meaning of a large body of input text \citep{rush-etal-2015-neural}. Typically, the input text consists of a continuous information flow, such as a story or an article. However, text data from sources like surveys, product reviews, and opinion boards (e.g., Twitter) exhibit a non-continuous information flow, where individual units of text, such as tweets, are not connected to one another. This type of data is common in the fields of education~\citep{zhao2021targeted, zhao2021end} and human resources~\citep{xu-etal-2024-hr}. However, it presents a challenge to text summarization.
    
\paragraph{Earlier Approaches}
    A common approach to summarizing such non-continuous data uses automatic summarization frameworks \citep{coavoux-etal-2019-unsupervised,10.1145/3448015}. This approach first clusters topics~\citep{xu-etal-2023-detime, xu-etal-2023-vontss, xu2023kdstm, xu2023s2vntm} and then summarizes. We call this approach the two-step baseline. The main problem with this method is the summarization model's dependency on topic clustering. The quality of summary from the summarization model depends on the quality of clusters generated by topic clustering and it is difficult to improve and guarantee the homogeneity of each cluster.
    Reinforcement Learning method is used in \citep{supert} to select salient sentences from multiple documents and generate a summary, but the input documents are all related to one topic, which isn’t the case in text data.
    In Aspect Based Summarization \citep{aspectBasedSummarization1,aspectBasedSummarization2}, the aspects are a known and are provided to the model which is not the case in text data.
    Figure \ref{fig:JADS_task} shows JADS compares to other approaches.

    Aspect detection \citep{simpleContrastiveAD} typically involves identifying aspects and/ mapping them to predefined aspects of interest. These aspects are usually represented by 1-3 word phrases.

\paragraph{ADS: Aspect discovery and summarization}
    In our case, Aspect discovery and summarization, involves identifying aspects or topics in a text that are not be pre-defined or limited to a fixed set of keywords. Instead of focusing on individual words or phrases, the goal is to identify key themes that emerge from the text and provide a summary of the main points being discussed. These summaries may not be represented by single words or phrases, but rather by longer, more descriptive phrases or sentences which maybe combinations of several topics (themes).

\paragraph{JADS: Joint Aspect discovery and summarization}

    To tackle the challenges posed by non-continuous text data summarization, we introduce the Joint Aspect Discovery and Summarization (JADS) approach. JADS builds on the ADS. It is a more flexible and exploratory approach which identifies key topics and themes that emerge from a given text.

JADS integrates aspect discovery and summarization tasks into a joint modeling framework. The key advantage of this approach is that the entire process becomes fully differentiable, addressing the issue of poor error propagation in multi-step processes commonly seen in automatic summarization frameworks. By jointly modeling aspect discovery and summarization, the JADS model learns to perform clustering and summarization simultaneously during training. Errors in the summarization process can flow back to the clustering step and ultimately to the inputs. This end-to-end differentiability allows for improved learning and optimization across both aspect discovery and summarization tasks.
        
    JADS can generate multiple summaries for multiple topics in the input documents with no explicit information about aspects, addressing the limitations of aspect-based summarization.
    
    In addition to introducing JADS, we propose a method for creating large-scale ADS datasets for text summarization based on existing summarization datasets. We use the Wikipedia dataset (CITE) for pre-training and use the CNN/Daily Mail (CNN/DM) dataset \citep{10.5555/2969239.2969428} for testing and experimentation. We hope that this data set creation method will encourage more research into joint topic summarization.

    In particular, our work makes the following contributions:
    \begin{itemize}
        \item We introduce Aspect Discovery and Summarization (ADS) task for text data.
        \item We propose a self-supervised method for creating large-scale ADS datasets based on existing summarization datasets.
        \item We present a novel method JADS which can jointly generate summaries for various aspects in text content. Our experiments show that JADS can achieve better performance than the baseline two-step method.
        \item We demonstrate that pretraining JADS on large wikipedia dataset (ADS converted) increases performance and stability of the model.
        \item we show JADS embeddings exhibit superior clustering capabilities.
        \item Human evaluation shows the method is semantically align with the ground truth and is factual.
    \end{itemize}
    
    In Section \ref{ads}, we provide an overview of ADS, JADS, self-supervised training, and pre-training. The details of baseline methods can be found in section \ref{sec_baseline}. We explain the evaluation methods used in section \ref{Sct::evaluation}. Section \ref{Sct::results} covers the results and ablation studies. Appendix \ref{relatedWork} contains additional ablation studies, more related work, experimental settings, and more.

\section{Aspect Discovery and Summarization}
\label{ads}

    In this section, we formalize the ADS task for text data and present our models. Given a collection of input text sentences $X$, our model produces a set of summaries $S$. The input $X$ has the following properties. 1) The sentences in $X$ contain an unknown number of topics;  2) sentences in $X$ have non-continuous information flow. In other words, the sentences are not organized by topic. Regarding the output $S$, we note the following. (1) Each topic in $X$ should have a summary in $S$. 2) each summary provides condensed yet salient information on the topic. 

There are two parts to building a model for the joint task: model structure and creation of pretraining data and pretraining. The following subsections discuss these aspects.

\subsection{JADS}
    Our model builds on the Longformer \citep{Beltagy2020Longformer}, a transformer-based encoder-decode model. We use the Longformer because it can fit in longer documents in the GPU memory and provides a good balance for time and memory complexities. Differing from traditional document summarization task, in ADS, there can be a set of summaries $S$ for $X$. To generate summary $S$ for $X$, $n$ sub-summaries are concatenated with a [SEP] token shown as follows:
    \begin{equation}
    \label{eq1}
        S = S_1\ [SEP]\ S_2\ [SEP],... [SEP]\ S_n
    \end{equation}
    where $S$ is a concatenated summary, $S_i$ is one of the sub-summary. With the cross-entropy used for fine-tuning as follows:
    \begin{equation}
    \label{eq2}
        \mathcal{L} = -\frac{1}{T} \sum_{t=1}^T\log P(w_t)
    \end{equation}
    where $T$ is the length of the concatenated summary and $w_t$ is the \textit{t}th token in the summary, the Longformer model is guided to learn to predict the separate token automatically.

    \begin{table}[!htb]
        \small
        \centering
        \begin{tabular}{llll}
        \hline
                                     $K$   & 2    & 3     & 4  \\ \hline
        Len of content           & 1,381(476) & 2,072(586) & 2,763(674) \\
        Len of summary           & 104(30)    & 157(37)     & 210(43)     \\
        \# of training          & 143,556    & 95,704     &71,778  \\
        \# of evaluation        & 6,684    & 4,456     & 3,342  \\
        \# of testing           & 5,745    & 3,830     & 2,872  \\  \hline
        \end{tabular}
        \caption{The statistics of our new datasets with different summary number (2,3, and 4). The standard deviations of average length of content and summary are reported in the bracket}
        \label{tab:datasets}
    \end{table}

\subsection{Self-supervised Training}
    \label{Sct::Dataset}
    To train and evaluate our method, we require an ADS dataset. Several summarization datasets comprising multifaceted documents have been proposed recently \citep{cohan-etal-2018-discourse,DBLP:conf/iclr/LiuSPGSKS18,aspectBasedSummarization2}. However, these datasets do not satisfy the ADS requirements mentioned in Section~\ref{ads}. 
    
    It is hard to create a large labeled dataset for this problem through human labeling. Hence, we propose a new approach to create a dataset for topic discovery and summarization. We create an ADS dataset leveraging Wikipedia (for pretraining) and CNN/Daily Mail (CNN/DM) datasets for fine-tuning.
    Our dataset is a set $\mathcal{D}$ of data points $d=(x,y)$, where $x$ is the text content, and y is a collection of summaries for all topics in $x$. 

    To construct $x$, and $y$ pairs, we sample $K$ article-summary pairs from the underlying dataset (Wikipedia or CNN/DM). $K$ is the number of topics/aspects.  We pick the text from the articles and aggregate them to get $x$. $y$ is the concatenation of summaries of the articles, in random order and separated by [SEP]. The algorithms used for sampling and aggregating are described in subsections below. For CNN/DM dataset, we use it's 'highlights' as summary and 'article' as article.

    In our experiments, we experiment with various summary numbers with 2, 3, and 4. For $k>4$, the capacity of our infrastructure is exceeded i.e. GPU gets out of memory for $K>4$ with Longformer model. In Table \ref{tab:datasets}, we show the average length of text content and summaries of our datasets for different summary number. We also show the number of samples for training, evaluation, and testing.
    
    For self-supervised training, each input $x$ is a collection of sentences belonging to $k$ document. We need to select $k$ documents from the corpus and order the sentences in them in some manner. The following subsections describe article selection and sentence ordering.
    
    \subsubsection{Article Selection}
    \label{Sampling}
    To sample $K$ pairs, we use two methods: Random sampling and \selapp. Random sampling is randomly choosing $K$ pairs from corpus and \selapp\ is choosing $K$ pairs whose summaries are semantically most distant to each other.
     Algorithm for \selapp\  is described in Algorithm \ref{alg1}. 
    
    We use the fine-tuned model ‘all-MiniLM-L6-v2’ \citep{wang2020minilm} for the ‘sentence\_embedding’ function. Table \ref{tab:sampling_method1} in the Appendix shows the comparative performance of the two methods for $K=2$. Based on the results, there is no significant difference in random sampling and \selapp\  and so we use random in rest of the experiments.

\subsubsection{Sentence Ordering}
\label{Aggregating}
To determine the order of sentences in $x$, we try three methods: inorder or no shuffling, in-sample, and cross-sample.

\begin{description}
\item[no-shuffling:] There is no shuffling of sentences, we concatenate articles in $x$ in order, so the text content is all sentences in sequence (no-shuffling).
\item[in-shuffling:] We randomly shuffle the sentences in each article before concatenating them (in-sample-shuffling). The relationship between sentences will be weak, though the sentences from the same article are still together but in different order.
\item[cross-shuffling:] We first concatenate all sentences of all articles in $x$ and then randomly shuffle all sentences (cross-sample-shuffling), which can actually simulate the of text content for ADS by breaking continuous information flow.
\end{description}

    To aggregate $y$ from $K$ pairs, we concatenate summaries using SEP token, as shown in Equation (\ref{eq1}).

\subsection{Pre-training}
\label{pretrainAppendix}
    We pre-train JADS on large Wikipedia dataset.
    We use self-supervised training described in section \ref{Sct::Dataset} to create dataset. Specifically, we use titles as summary $y$ and texts of article as the input $x$. We create a $K=4$ summary ADS dataset using the $x$ and $y$ article summary pairs from this dataset and pre-train the model on this dataset. The dataset configuration is “wikipedia”, “20220301.en”, revision="master" contains training set: 1,453,201 and validation set: 161,468. We pre-train the model for 4 epochs. Then we use this pre-trained model as a starting point and train it on “random-choosing\_cross-sample-shuffling” dataset and call this method 'JADS-pre' in results.

\section{Baselines}
\label{sec_baseline}
This section shows the details of various baseline methods used in our study.

\subsection{Two-step method}
        For the ADS, we introduce one straight forward baseline, which is first to cluster the sentences in the content, and, for each cluster, we generate one summary based on the sentences in that cluster.

        For the step of clustering, we utilize BERTopic \citep{BERTopic} which is one of the most competitive topic modeling methods to generate the specific number (summary number) of sentences clusters based on the text content. BERTopic is a topic modeling technique that leverages transformers to create dense clusters. BERTopic generates document embedding with pre-trained transformer-based language models based on the Sentence-BERT framework \citep{reimers-gurevych-2019-sentence}, then reduces embedding dimension with UMAP model \citep{mcinnes2018umap-software}, and clusters the reduced embeddings with HDBSCAN \citep{mcinnes2017hdbscan}. 
        
        In our experiments, we keep the default hyperparameters for the BERTopic model except for the 'embedding model', 'nr\_topics' and 'random\_seed'. We use ‘allenai/led-base-16384’ for embedding model in BERTopic since the default model BertTopic uses i.e. ‘all-MiniLM-L6-v2’ has input sequence length limit of 128 which is much lower than the input length of content shown in Table \ref{tab:datasets} whereas Longformer can process up to 16k tokens. Since the BERTopic model might generate more clusters than the summary number, we explore multiple methods to resolve this which are discussed in Appendix  \ref{Sct:two-step clustering reduction} and choose ‘least’ in the rest of the experiments since it's the default method of BERTopic. We keep the 'min\_topic\_size' to its default value 10 since it much is greater than $K$. In umap\_model, we change the 'transform\_seed' and 'random\_state' to keep them same as our 'random\_seeds' which is discussed in Appendix \ref{ExperimentalSettings}. There are no changes made to HDBSCAN model in BERTopic.

        For the step of summary generation, we fine-tune a Longformer model on the original CNN/DM dataset which contains 1-to-1 article-summary pair and use it to generate summary based on the given clustered sentences. The length of the input tokens is 1024, and the length of the output tokens is 128.
        
\subsection{Two-step-pre method}
        For this method, we use replace the embedding model used in two-step method i.e. ‘allenai/led-base-16384’ with the model pre-trained on wikipedia dataset. The pretraining is explained in Section \ref{pretrainAppendix}. We call this baseline as two-step-pre.The purpose of this baseline is to determine whether pre-training enhances the clusterability of embeddings.

\subsection{Two-step-VAR method}
        For this method, we use replace the embedding model used in two-step method i.e. ‘allenai/led-base-16384’ with the model trained on variable summary number dataset (random-choosing\_cross-shuffling) discussed in section \ref{scalability}. We call this baseline as two-step-Var. This baseline will show if clusterability of embeddings is improved after being fine-tuned on variable dataset.
    
\subsection{Two-step-K method}
        For this method, we use replace the embedding model used in two-step method i.e. ‘allenai/led-base-16384’ with the model trained on summary number K dataset (random-choosing\_cross-shuffling). We call this baseline as two-step-Var. This baseline will show if clusterability of embeddings is improved after being fine-tuned on summary number K dataset.

\section{Evaluation}
    \label{Sct::evaluation}
    To compare the performance of JADS and the multi-step baseline, we use The Rouge metric \citep{lin-2004-rouge} for evaluating the quality of produced summaries. In our experiments, we rely on Rouge 1, Rouge 2, Rouge L, and Rouge L Sum. Rouge 1 and Rouge 2 are calculated based on the common n-gram (1-gram for Rouge 1 and bigram for the Rouge 2) in the generated summary and reference. Rouge L and Rouge L Sum are calculated based on the longest common sequence in the generated summary and reference. 

    For a given $x$, $y$ in ADS dataset, the model can generate $M$ summaries while the number of ground truth summaries is $K$. When the model generates more or fewer summaries than the ground truth, we use cluster mapping techniques to match the clusters. They are described in Appendix \ref{cluster_mapping}. To ensure $M=K$, we take the following post-process steps: For JADS, we select 'closest' method of three discussed in detail in Appendix \ref{Sct:JADS clustering reduction}. For the two-step baseline, we ensure $M \leq K$ when generating clusters which is discussed in Appendix \ref{Sct:two-step clustering reduction}. For both methods, if $M<K$, we will add empty summaries to the generated summaries to ensure that the number of generated summaries is equal to the number of references.

    Since the order of generated summaries is not important for this task, for each sample, we pair the generated summary $M_i$ with $K_j$ where $Mi\in M$ and $K_j \in K$ based on the similarity map generated using RougeSum (shown in equation \ref{eq3}). We select the reference $K_i$ for $M_i$ which has the highest RougeSum and if there is a tie, we select the first one.


\begin{equation}
\label{eq3}
\begin{aligned}
    \text{RougeSum}(M_i,K_j) &=  \text{Rouge1}(M_i,K_j) \\
    &+ \text{Rouge2}(M_i,K_j) \\
    &+ \text{RougeL}(M_i,K_j) \\
    &+ \text{RougeLSum}(M_i,K_j)
\end{aligned}
\end{equation}

\begin{table*}[!ht]
\renewcommand{\arraystretch}{1.5} 
\LARGE
\hspace*{1cm} 
\scalebox{0.40}{
\begin{tabular}{|l|lll|lll|lll|}
\hline
        $K$     & \multicolumn{3}{c|}{2}                                                    & \multicolumn{3}{c|}{3}                                                & \multicolumn{3}{c|}{4}                           \\ \hline
                & JADS        & JADS-pre               & two-step        & JADS        & JADS-pre            & two-step      & JADS        & JADS-pre            & two-step           \\ \hline
    Rouge 1     & 38.51(0.11)  &\textbf{39.44(0.10)}   & 34.93(0.15)     & 37.33(0.09) &\textbf{38.74(0.08)} & 26.98(0.14)   & 36.38(0.17) &\textbf{37.30(0.04)} & 23.27(0.13)       \\
    Rouge 2     & 16.38(0.10)  &\textbf{16.96(0.07)}   & 13.14(0.17)     & 15.61(0.11) &\textbf{16.47(0.09)} & 10.01(0.06)   & 14.98(0.15) &\textbf{15.61(0.06)} & 8.46(0.10)      \\
    Rouge L     & 26.64(0.13)  &\textbf{26.83(0.06)}   & 22.76(0.14)     & 25.94(0.02) &\textbf{26.31(0.09)} & 17.55(0.04)   & 25.24(0.19) &\textbf{25.38(0.04)} & 15.12(0.11)        \\
    Rouge L Sum & 35.77(0.10)  &\textbf{36.46(0.10)}   & 32.37(0.19)     & 34.68(0.07) &\textbf{35.80(0.06)} & 25.05(0.14)   & 33.77(0.17) &\textbf{34.36(0.03)} & 21.60(0.14)        \\ \hline
    $\mu\Delta_{\text{{two-step}}}$ & 3.52 & \textbf{4.12} & 0   & 8.49 & \textbf{9.43} & 0  & 10.48 & \textbf{10.93} & 0  \\ \hline

\end{tabular}
}
\caption{The comparison of JADS (ours), JADS-pre (ours) and two-step (baseline) and two-step-pre (baseline) methods with Rouge 1, Rouge 2, Rouge L and Rouge L Sum. The experiments are conducted with ``random-choosing\_cross-sample-shuffling'' dataset. The standard deviation is reported in the bracket. $\mu\Delta_{\text{{two-step}}}$ \text{is mean of difference of rouge metrics of two-step method and other methods.}}
\label{tab:ours-baseline comparision}
\end{table*}

\begin{table*}[!ht]
\renewcommand{\arraystretch}{1.5} 
\LARGE
\scalebox{0.35}{
\begin{tabular}{|l|llll|llll|llll|}
\hline
$K$ & \multicolumn{4}{c|}{2} & \multicolumn{4}{c|}{3} & \multicolumn{4}{c|}{4} \\ \hline
\multirow{2}{*}{} & two-step & two-step-pre & two-step-JADS-var & two-step-JADS-2 & two-step & two-step-pre & two-step-JADS-var & two-step-JADS-3 & two-step & two-step-pre & two-step-JADS-var & two-step-JADS-4 \\ 
                  &          &              &                   &                 &          &              &                   &                 &          &              &                   &                 \\ \hline
Rouge 1           & 34.93(0.15) & 33.29(0.05)  & \textbf{35.57(0.09)} & 35.21(0.18)    & 26.98(0.14) & 24.92(0.06) & \textbf{27.86(0.15)} & 27.79(0.07) & 23.27(0.13)  & 20.60(0.23)  & \textbf{24.62(0.16)} & 24.90(0.15) \\
Rouge 2           & 13.14(0.17) & 12.09(0.08)  & \textbf{13.59(0.09)} & 13.32(0.15)    & 10.01(0.06) & 8.93(0.10)  & \textbf{10.44(0.09)} & 10.42(0.03) & 8.46(0.10)   & 7.24(0.17)   & \textbf{9.12(0.07)}  & 9.22(0.14) \\
Rouge L           & 22.76(0.14) & 21.64(0.07)  & \textbf{23.22(0.11)} & 22.97(0.15)    & 17.55(0.04) & 16.19(0.01) & \textbf{18.10(0.13)} & 18.09(0.04) & 15.12(0.11)  & 13.38(0.19)  & \textbf{16.04(0.09)} & 16.19(0.11) \\
Rouge L Sum       & 32.37(0.19) & 30.81(0.06)  & \textbf{32.99(0.09)} & 32.64(0.20)    & 25.05(0.14) & 23.10(0.08) & \textbf{25.85(0.13)} & 25.78(0.06) & 21.60(0.14)  & 19.11(0.25)  & \textbf{22.86(0.13)} & 23.11(0.15) \\ \hline
$\mu\Delta_{\text{two-step}}$ & 0 & -1.34 & \textbf{0.54} & 0.23 & 0 & -1.61 & \textbf{0.66} & 0.62 & 0 & -2.03 & \textbf{1.04} & 1.24 \\ \hline
\end{tabular}
}
\caption{The comparison of two-step baseline with embeddings from various models. Pre is pretrained, JADS-var is JADS model trained on variable dataset, JADS-N is JADS model trained on K=N dataset. The standard deviation is reported in the bracket. $\mu\Delta_{\text{{two-step}}}$ is the mean of difference of rouge metrics of two-step method and other methods.}
\label{tab:all-baseline comparision}
\end{table*}

\section{Results}
\label{Sct::results}

        To compare the performance of our JADS methods and the baseline methods, we conduct experiments on the ``random-choosing\_cross-sample-shuffling’’ dataset with different summary numbers. Table \ref{tab:ours-baseline comparision} shows the performance of our JADS methods and the two-step baselines. We can observe that our JADS method achieves better performance than the two-step baseline. With the increase in summary number, i.e. when there are more clusters  JADS is expected to outperform the two-step method for ADS. JADS can learn from both the steps, i.e. clustering and summarization, and errors from summarization flows to the inherent clustering step. We compare the clustering performance of JADS and two-step baseline in ablation studies in Section \ref{clusterPerformanceAnalysis}. JADS-pre pre-trained on wikipedia dataset shows higher performance than JADS method which shows that pre-training helps the JADS model.
        
        To understand whether the non-continuous information flow of sentences in data is more difficult for model to learn, we compare the performance of our model trained with different shuffling method on the ``random-choosing’’ dataset. In Table \ref{tab:performance with different shuffling} in Appendix, we can observe that, for Rouge 1, 2 and L, the ``no-shuffling’’ achieves the best performance, and ``cross-sample-shuffling’’ achieves the worst performance. This is in accordance with our intuition and proves that the non-continuous information flow increases the difficulty of ADS task. However, for Rouge L Sum, ``in-sample-shuffling’’ achieves the best performance. Compared to Rouge-N which computes N-gram scores, Rouge L computes longest common subsequence (LCS) between two pieces of text and ignores newlines. In Rouge L Sum, newlines are treated as sentence boundaries and LCS is computed between each pair of sentences. To calculate Rouge L Sum, we add newline between summaries and this is the reason Rouge L Sum of in-sampling-shuffling is higher.

        To measure the impact of article selection methods, specifically random sampling and \selapp\  methods, we compare the performance of JADS on ``cross-sample-shuffling’’ dataset with $K=2$. The results are shown in Table \ref{tab:sampling_method1}, in which we can observe that the performance for the \selapp{} method is slightly better than the random method. The difference is not significant, which proves that our method is not sensitive to the similarity of the summary. 

        We conducted Human Evaluation shown in Appendix \ref{humanEval}. In Appendix \ref{Sct:Examples}, we show the examples of good and bad performance for our methods.

\subsection{Ablation}\label{ablationAppendix}
    In this section, we show various ablation studies conducted to test the scalability of the algorithm, clustering performance analysis between baseline and JADS, effect of cluster mapping algorithms, effect of pre-training and effect of using pre-trained model embeddings in baseline. More ablation studies are added in Appendix \ref{moreAblationStudies} due to limited space.

    \subsubsection{Scalability of the Summary Number}
    \label{scalability}
\begin{figure}
        
        \centering
        \includegraphics[scale=0.48]{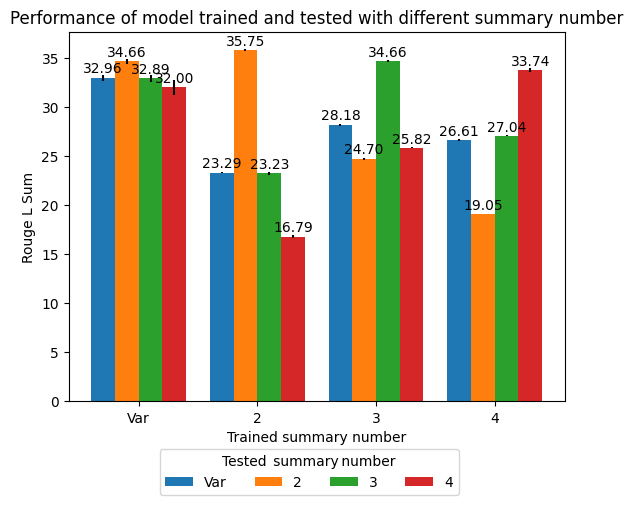}
        \caption{The performance of models trained and tested with various summary number}
        \label{fig:summary_number}
        \end{figure}

        To test JADS on real life text data, we experiment with the scalability of summary number. Here, we test if a model trained with one summary number is able to predict any number of summaries (e.g. a model fine-tuned with three summaries should be able to predict 2 or 4 summaries). Therefore, we test the model on a different summary number with ``random-choosing\_cross-sample-shuffling’’ dataset. We also test if our method generates variable summary number outputs, therefore, we combine ``K\_random-choosing\_cross-sample-shuffling’’ (where $K$ = 2,3, and 4) dataset together for fine-tuning and testing. To ensure the number of generated summary number is smaller than ground truth summary number, instead of iteratively merging the closest two summaries (described in Section \ref{Sct::evaluation}), we pad empty summaries to the ground truth data to simulate the situation in which we do not know the summary number. The results for both experiments are shown in Figure \ref{fig:summary_number}. 
        
        In Figure \ref{fig:summary_number}, we can observe that when given a trained summary number, the model achieves the best performance on the same tested summary number. This phenomenon shows that our approach will lead the model to generate a fixed number of summaries. By observing the performance of the model trained with variable summary number dataset (Var), we can observe that its performance is the most consistent when tested with different summary numbers. This proves that, by using a dataset with various summary numbers, the model’s bias to generate a fixed number of summaries can be partly addressed. 
        
        Since the datasets for k=2,3,4 are merged together in Var, articles are reused in various input pairs. Results on Var dataset show that model is not over-fitting to the input text or the topics for eg. if input has topic/article 'football', k=4 or 3.

    \subsubsection{Clustering Performance Analysis}
        \label{clusterPerformanceAnalysis}
        Compared with the two-step baseline, our JADS method achieves significantly better performance. Experiments in Section \ref{Sct::results} shows that the low performance of the two-step method is mainly because of the clustering step. To understand the quality of clustering for our JADS method in the aspect of: 1) number of generated clusters; 2) the similarity of the generated and ground truth clusters. we conduct experiments on the random-choosing\_cross-sample-shuffling datasets with different summary number (2,3,4, and variable summary numbers). For the number of generated clusters, we regard each generated sub-summary as one cluster. For the aspect of homogeneity of generated clusters, for each sentence in one sample, we use the BERTScore to understand the most relevant generated summary. Then we regard all the sentences highly relevant to one generated summary as one class. Macro-F1 is used for measuring the performance of clustering. To map the generated labels and ground truth labels, we exhaustively test all mapping relationships and regard the one with the highest F1 as the correct mapping relationship. The results are shown in Table \ref{tab:clustering analysis}. For the baseline two-step method, for variable summary number, we set nr\_topics to ‘None’, where no cluster reduction happens i.e. all the clusters produced by HDBSCAN are used directly. 
        
\begin{table}[!ht]
\centering
\scalebox{0.8}{
\begin{tabular}{l|lll}
\hline
$K$            & Method   & \# of clusters       & F1                   \\ \hline
\multirow{2}{*}{$2$}   & JADS     & \textbf{2.00(0.01)} & 0.86 (0.13)        \\
               & Two-step & 2.04(0.19)          & \textbf{0.89(0.15)} \\ \hline
\multirow{2}{*}{$3$}   & JADS     & \textbf{3.00(0.03)} & \textbf{0.79(0.14)} \\
               & Two-step & 2.43(0.56)          & 0.64(0.21)          \\ \hline
\multirow{2}{*}{$4$}   & JADS     & \textbf{4.00(0.02)} & \textbf{0.73(0.13)} \\
               & Two-step & 2.88(0.88)          & 0.54(0.21)          \\ \hline
\multirow{2}{*}{Var} & JADS     & \textbf{2.95(0.81)} & \textbf{0.76(0.16)} \\
               & Two-step & 2.54(0.83)          & 0.63(0.29)      \\ \hline 
\end{tabular}}
\caption{The analysis of the clustering quality for JADS and two-step baselines. ``Number'':summary number,  ``Var'':various number dataset. The reported ``\# of clusters'' and std in the bracket is the mean of three random seeds. The average number of summaries for ``Var'' is 3.}
\label{tab:clustering analysis}
\end{table}
By observing the number of clusters in Table \ref{tab:clustering analysis}, we know that our method, JADS is more likely to generate the correct number of clusters, especially when the summary number is high. When focusing on the F1 score, we can observe that our JADS achieves better performance than the two-step baseline. This two observations are in accordance with the performance shown in Table \ref{tab:ours-baseline comparision}. It also indirectly proves that our model archives better performance when the summary number is large because of better clustering.
        We compare k=8 on the longformer and longT5 \citep{longt5} models which is maximum what longformer supports in Appendix \ref{sct:k=8}.

\subsubsection{JADS Embeddings for two-step baseline}    

    In this section, we compare embeddings from original longformer ('allenai/led-base-16384'), two-step-pre (JADS pre-trained),JADS-var model (JADS trained on variable K dataset.) and JADS-k model (JADS trained on summary number k dataset), , with the two-step baseline. The experiments are conducted with ``random-choosing\_cross-sample-shuffling'' dataset.
    
    From table \ref{tab:all-baseline comparision}, it can be observed that two-step method has degraded performance when using embeddings from JADS pre-trained model compared to original longformer model. This can be due to the fact that original longformer has weights from BART \citep{bart} which was pre-trained on much larger and diverse corpus. The pre-training included multiple tasks as opposed to single task in ADS. But the opposite is true when JADS is finetuned using this pre-trained model as starting point, in which it increases the model performance as seen in table \ref{tab:ours-baseline comparision}. Similar observation can be made from Table \ref{tab:baseline-pretranied-embeddings} where experiments are conducted on ``random-choosing\_no-shuffling'' dataset.
    
    Two-step-JADS-var and two-step-JADS-k  consistently have higher performance over the original embeddings in two-step method. The performance gap widens as the K increases. It is interesting to see that the two-step-JADS-var model's embeddings match the performance of two-step-JADS-k model, this further proves that Var model as well as it's embeddings are scalable for various summary numbers. 

\subsubsection{Sentence Ordering and Upper Limit Results}

    To get the upper limit of performance of JADS on the dataset, we  test summary number 1 on 'random-choosing\_no-shuffling' dataset. This means that only one article-summary pair is given, and this becomes a regular summarization task where no clustering is involved. The results are shown in Table \ref{tab:performance with summary number} in Appendix. We observe that there is a small performance drop from $K=1$ to $K>1$ but the results are almost same for K=2,3,4. This shows that addition of clustering induces a slight performance drop, but it remains almost the same for $K>1$.

\subsubsection{Cluster Alignment and Article Selection Results} \label{sec:align}

    In this section, we report the results of cluster mapping and article selection on baseline and JADS. 
    
    Table~\ref{tab:two-step cluster reduction} presents the results of cluster mapping for the baseline method. We observe that there is no significant difference in the results between the different mapping methods, particularly when the $K$ is greater than 2.

    Similarly, Table~\ref{tab:model_reduction_method} displays the results of cluster mapping for JADS. Similar to the baseline, we find no significant difference between the mapping methods used.
    
    In Table~\ref{tab:sampling_method1}, we provide the results for different article selection methods in JADS with $K$=2. Our analysis indicates that there is no significant difference in the results across these different article selection methods.

\section{Conclusion, Limitations and Future Work}
    In this paper, we present ADS task in which a model generates one summary for each of the aspects or topics in the text data. Sizable datasets for ADS task does not exist and are expensive to create. We propose a scalable self-supervised method that leverages existing datasets and makes it suitable to our task. In our experiments, we show that all the following three factors: the summary number, sampling method, and shuffling method influences the difficulty of the text summarization. These experiments provide paths for generating better ADS datasets in the future.

    We apply encoder-decoder longformer model for abstractive summarization task and provide a baseline for this task. Our results prove that, compared with the baseline method, our method can achieve better performance by jointly modeling clustering and summarization steps. We also show that, differing from aspect-based summarization, our method is not sensitive to the number of topics/aspects in the summary when trained with a dataset comprising variable summary numbers, which ensures the scalability of our method.

    We pre-trained JADS on transformed ADS Wikipedia dataset and show that, with pretraining, there is an improvement in model performance and stability.

    \paragraph{Limitations}
    In ADS data, there can be an unknown number of topics. From our experiments, we saw JADS generates a specific number of summaries when trained on dataset with a fixed number of topics, i.e. it is biased to generate a fixed number ($K$) of summaries used in train set. We analyse the scalability of this model on various summary numbers but the performance drops when the number of topics in the test is not same as that in the training set. We partly address it by using a variable  number of topics in the dataset for training.
    In this paper, we have not shown scalability of the JADS method for $K>4$ due to GPU memory limitations and the ADS data might contain unknown number of topics, therefore this is a major limitation.

\bibliography{main.bib}
\bibliographystyle{acl_natbib}

\clearpage
\appendix

\textbf{\large Appendix}

\section{Human Evaluation}
\label{humanEval}
In our research, we conducted a human evaluation of the generated summaries. For this purpose, we used Amazon Mechanical Turk, where annotators were tasked with answering various questions given ground truth and model generated summaries. We randomly sampled 500 summaries and evaluated on various questions. Each item wass assigned to 3 annotators and we have shown all the labels with ground truth confidence > 60\%.

\subsection{Q1}

\begin{figure}
    \centering
    \includegraphics[width=\linewidth]{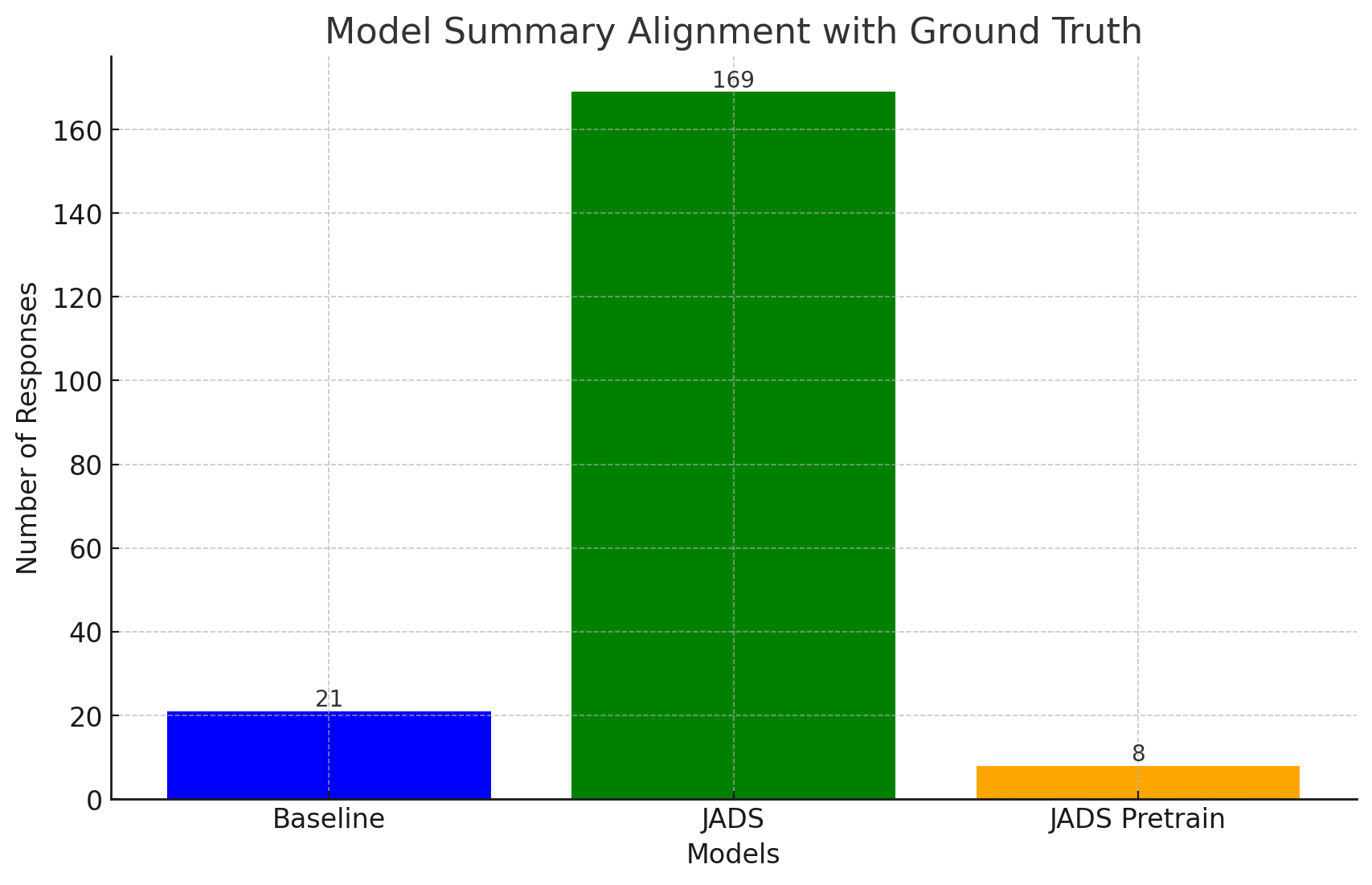}
    \caption{Distribution of responses indicating which model's summary is more aligned with the ground truth.}
    \label{fig:jads_aligned_eval}
\end{figure}

On asking annotators which model's summary is more aligned with the ground truth, majority of the labelers preferred JADS over baseline. The resutls are shown in \ref{fig:jads_aligned_eval}.

\subsection{Q2}

\begin{figure}
    \centering
    \includegraphics[width=\linewidth]{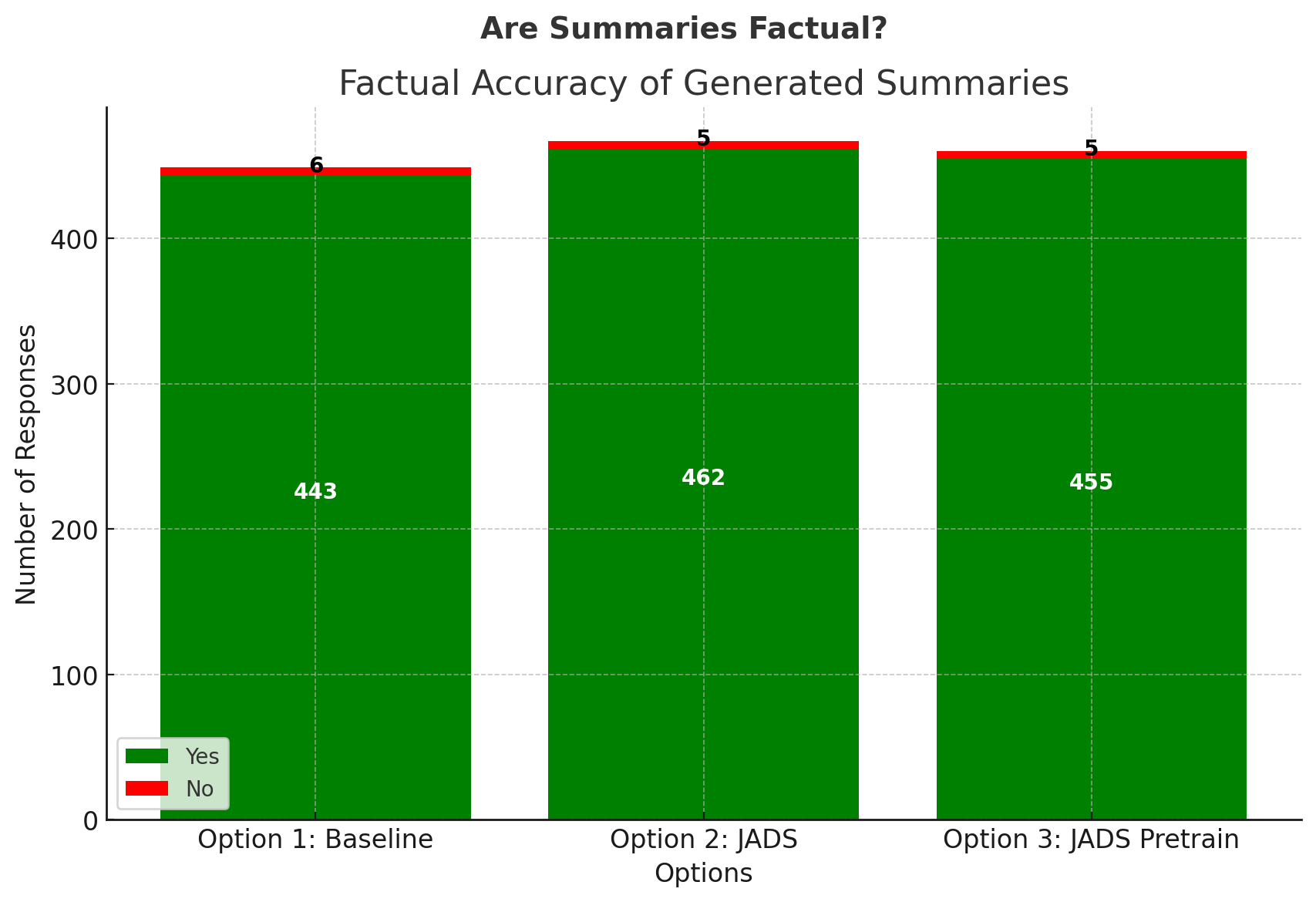}
    \caption{Assessment of factual accuracy in generated summaries across different options. 'Yes' indicates factual alignment, while 'No' indicates a deviation from factual content.}
    \label{fig:factual_accuracy}
\end{figure}

On asking annotators if generated summary is factual based on ground truth, these were the responses. We can see that most of the labelers found summaries from JADS models factually correct in \ref{fig:factual_accuracy}.

\section{More Models and larger $K$}
\label{sct:k=8}

We test the longformer model to it's extreme with it's maximum output length to 1024 i.e. for k=8. The scores are shown below. To compare with a different model, we used longT5 \citep{longt5} with the same training settings. We can see that the scores are in line with our hypothesis and remain steady even for higher topics.
The baseline crashes for $k=8$ in our experiments so their scores are not reported.
\subsection{Longformer model}

rouge1:33.488\\ 
rouge2:13.6001\\
rougeL:23.0177\\
rougeLsum:31.0099\\

\subsection{LongT5}
rouge1 30.5751 \\
rouge2 11.8564 \\
rougeL 21.1501 \\
rougeLsum 28.0818 \\


    \section{Future Work}
    
    In future work, we will explore the decoder-only model such as GPT to compare its performance with Longformer for ADS task. We will explore scalability of the model trained on one domain and test it on a different domain to test it's cross-domain capabilities. When $K>4$, the GPU memory limit is reached for Longformer model, we will analyze various methods and models such as RWKV, MPT-7b (84k input length) and fine-tuning techniques such as LoRA \citep{lora} to address this limitation in our future work.

\section{Related Work}
\label{relatedWork}
In Aspect Based Summarization \citep{aspectBasedSummarization2}, input texts along with aspect for which summary is to be generated is given to the summarization model. The aspects are small predefined set and therefore this method does not generalize well on other diverse aspects outside the predefined set. Authors in \citep{aspectBasedSummarization1} addresses this problem and proposes a new technique which summarizes arbitrary aspects relevant to the document. This method increases the predefined set of aspects but is still limited. The problem here is there exists a predefined set of aspects, whereas in text data, there can be many diverse aspects, i.e. it’s an open world.

In Extreme Summarization \citep{extremeSummarization1}, the input is a very long document containing several sentences, and the output is a one sentence summary. The input document contains paragraphs which describe themes around one topic. Sentences in these paragraphs are all connected to each other, i.e. information flow is continuous from sentence to sentence. The authors propose an abstractive summarization model based on convolutional neural networks, which is conditioned on topics from the dataset’s articles and can capture long range dependencies. In this work, the input and output is longer compared to \citep{extremeSummarization1} and besides that, nor the paragraphs nor the sentences have continuous information flow.

Automatic Text Summarization frameworks such as \citep{ldasummarization} and \citep{ldasummarization2} use semantic similarity based clustering and topic modeling using Latent Dirichlet Allocation (LDA) \citep{lda} to summarize large volumes of data in a multi-step process. Final summary highly depends on the cluster quality and here the error doesn’t propagate to the clustering stage. Our method generates summaries from input documents in one shot and the model is fully differentiable, which means the model can learn from errors in output. 

In \citep{supert}, authors train a reinforcement learning summarizer using their proposed unsupervised metric called SUPERT as rewards for the problem of multi-document summarization. The agent selects salient sentences from a document and then concatenates them to make a summary, which is extractive summarization. This method takes input documents which are all related to one topic and each of them have continuous information flow. In our work, neither the texts are all from one topic nor do they have continuous information flow. This method also comes with the overhead of optimizing parameters such as number of topics and threshold for document similarity.

\section{More Ablation Studies}
\label{moreAblationStudies}

\subsection{Cluster Mapping}
\label{cluster_mapping}
When the model generates more or fewer summaries than the ground truth, we use cluster mapping techniques to match the clusters. They are described in the following sub-sections. 
    \subsubsection{Two-step baseline: least, closest, longest, shortest}
    \label{Sct:two-step clustering reduction}
        For the two-step method, when the number of generated clusters is greater than the ground-truth summary number (e.g. generate 3 clusters while the ground-truth has 2 summaries), we follow the work of BERTopic \citep{BERTopic}, in which we iteratively, based on the c-TF-IDF, merge the least common clusters with its most similar one to reach the specific number (``least''). 
        We explore three more methods: 1) 'closest' in which we iteratively merge topics based on cosine similarity of their c-TF-IDF till we get $K$. 2) 'Longest' in which we select longest $K$ the clusters based on sum of sentence length and 3) 'Shortest' in which we select shortest $K$ cluster based on sum of sentence length. 
        
        From Table \ref{tab:two-step cluster reduction} in the Appendix, we can observe that all four methods perform very similar to each other. 'Longest' performs best for all values of $K$. 'Least' is second best for $K>2$, but for $K=2$ 'closest' performs better. 
        
        To further investigate this performance, we do the cluster quality analysis for these mapping methods in Table \ref{tab:clustering_analysis_two_step_methods} in Appendix. Based on the results we can observe that merging clusters reduces he homogeneity from 'Longest' method's 0.88 to 'least' method's 0.85 for $K=2$. For $K>2$, difference is less than 0.01. Similar is the case for number of clusters. The number of clusters and homogeneity performance here has direct impact on the summarization scores which are shown in Table \ref{tab:two-step cluster reduction}.
        
        The difference between all the methods not so significant and therefore we use BERTopic's default method 'least' in rest of our experiments.

    \subsubsection{JADS method: closest, longest, shortest}
    \label{Sct:JADS clustering reduction}
    In our experiments, we face the problem that JADS might generate more summaries than the ground truth summaries, especially for the model trained with variable summary number dataset (the combination of 2, 3, and 4 summaries). Therefore, we explore three methods merging summaries: 1) we iteratively, based on BERTScore, merge the closest summary until the specific number is reached (closest); 2) We choose the top $K$ longest summaries (longest); 3) we choose the top $K$ shortest summaries (``shortest'') where $k$ is the summary number.
    
    
    The performance of these three methods with combinations of summary number, sample choosing methods, and cross-sentence-shuffling method  is shown in Table \ref{tab:model_reduction_method} in the Appendix.

    As we can observe from Table \ref{tab:model_reduction_method}, the difference of these three methods is limited and not significant for the fixed summary number (2, 3, and 4). For the various summary, the ``closest'' method can achieve slightly better performance. Therefore, in our main experiments, we use the ``closest'' method.


\section{Experimental Settings}
\label{ExperimentalSettings}
    In this section, we talk about the settings for our experiments.
    
    We use 'allenai/led-base-16384' version of Longformer \citep{Beltagy2020Longformer} from huggingface for all the experiments. 
    \textbf{Input/Output Length} For our method, the Longformer we set input length to $K*1024$, and the output length to $K*128$. We set these two values because: 1) This ensures that model’s input and output length is more than average lengths in the dataset. Average lengths for dataset are shown in Table \ref{tab:datasets}); 2) the Longformer model will automatically pad the inputs to a multiple of the attention window, which is 1024 in our experiments. For the two-step baseline, since it only includes one article-summary pair, the input length is 1024 and the output length is 128. 
    
    For the two-step method’s first stage of clustering, we use BERTopic v 0.11.0 to run all the experiments with all default parameters except embedding\_model and 'nr\_topics' in which we use ‘allenai/led-base-16384’ pre-trained longformer. In addition to that, we change 'transform\_seed' and 'random\_state' in the umap\_model of BERTopic and set them to random\_seeds discussed below.
    
    \textbf{Random Seeds} For all experiments, we conduct three different random seeds (0, 10, 42). With different random seeds, we can reduce the influence of the random factor in our results. All results reported in Section \ref{Sct::results} are the mean score of three experiments.

\section{Algorithm} \label{algorithmAppendix}
In this section, the algorithm used for \selapp is shown where $S_i.summary$ denotes the summary of sample $S_i$.

\begin{algorithm} [!ht] 
\renewcommand{\algorithmicrequire}{\textbf{Input:}}
\renewcommand{\algorithmicensure}{\textbf{Output:}}
\caption{\selapp\ Selection} 
\label{alg1} 
	\begin{algorithmic}
		\REQUIRE  summary number K, the set of samples S
		\ENSURE The aggregated result $AR \gets \emptyset$
		\FORALL{$S_i \in S$}
		    \STATE $E_i \gets sentence\_embedding(S_{i}.summary)$ 
		\ENDFOR
		\STATE unused samples $US \gets S$
		\FOR{$i \in [1,|S|]$}
		    \IF{$S_i \notin US$} 
		        \STATE continue
		    \ENDIF
		    \STATE aggregated sample $AS \gets \{S_i\}$
		    \STATE $US \gets US - S_i$
		    \WHILE{$|AS|<K$ and $|US| \neq 0 $}
		        \STATE $S_c \gets \arg\max_{S_c \in US}(cosDist(S_c, AS))$
		        \STATE $AS \gets AS \cup \{S_c\}$
		        \STATE $US \gets US - S_c $
		    \ENDWHILE
		    \STATE $AR \gets AR \cup AS$
		\ENDFOR
	\end{algorithmic} 
\end{algorithm}

\section{Summarization Examples}
    \label{Sct:Examples}
    
    In this section, we show the examples of summary generated by various models on the ``random-choosing\_cross-sample-shuffling'' and $K=4$ test dataset. The order of summaries in the examples shown is random. 'label' column is used to denote the ground truth summary. Mean rouge sum shown at the end of the table shows mean of rouge\_mean of all the $K$ summaries in the example for that model. Rouge\_mean is the mean of Rouge1, Rouge2, RougeL and RougeLSum. Rouge\_mean is only used to select summarization examples.
    
    In Table \ref{tab:best_baseline}, example where baseline performs the best and generate summary for all four topics. Both JADS and baseline captures 4 topics and generates summary for them whereas pre-trained JADS misses the topic $K=3$. It produces two summaries for topic $K=0$.
    
    Table \ref{tab:JADS_best}, an example where JADS has highest rouge mean score of 50.96. Here baseline summary $K=3$ quality is poor and contains repeated sentences. pre-trained JADS performs even better than best performing example of JADS.
    
    Table \ref{tab:pre-trained_best} shows an example where pre-trained JADS has highest rouge mean score of 56.33. Summaries produced by JADS contains less information and baseline fails to produce 2 summaries here.
    
    Table \ref{tab:pre-trained_jads_bad} shows an example where pre-trained JADS has the lowest score. Even though all 3 methods produces 4 summaries, summaries from baseline and JADS are comparatively better than pre-trained JADS.

    Table \ref{tab:baseline_worst} shows an example where baseline performs worst. Here baseline's clustering produces only 2 clusters and therefore only 2 summaries are generated instead of 4. JADS and pre-trained JADS both produce 4 summaries.
    
    Table \ref{tab:jads_worst} shows an example where JADS has the lowest rouge mean score. Here JADS fails to produce 4 summaries and summary at $K=2$ contains repeated sentences and is of poor quality. Baseline and pre-trained JADS both produce 4 summaries.

    

\begin{table*}[!ht]
\begin{tabular}{l|ll|ll|ll}
\hline
         $K$       & \multicolumn{2}{c|}{2}                       & \multicolumn{2}{c|}{3}                                  & \multicolumn{2}{c}{4}              \\ \hline
                          & two-step             &two-step-pre    & two-step             &two-step-pre                      & two-step             &two-step-pre  \\
    Rouge 1               & \textbf{37.83(1.65)} & 36.98(0.10)    & \textbf{30.30(0.34)} & 28.05(0.77)                      & \textbf{26.68(0.08)} & 23.66(0.15)  \\
    Rouge 2               & \textbf{16.34(0.60)} & 15.35(0.15)    & \textbf{13.09(0.22)} & 11.67(0.06)                      & \textbf{11.44(0.03)} & 9.84(0.06)   \\
    Rouge L               & \textbf{26.05(1.11)} & 25.27(0.13)    & \textbf{20.84(0.26)} & 19.20(0.06)                      & \textbf{18.38(0.05)} & 16.22(0.10)  \\
    Rouge L Sum           & \textbf{33.72(3.73)} & 32.78(2.24)    & \textbf{28.30(0.31)} & 26.16(0.04)                      & \textbf{23.76(1.62)} & 21.03(1.59)  \\ \hline
\end{tabular}
        \caption{Baseline performance when using JADS wikipedia pre-trained model embeddings on random-sample\_no-shuffling dataset.}
\label{tab:baseline-pretranied-embeddings}
\end{table*}

    \begin{table*}[!ht]
        \centering
\begin{tabular}{|l|llllll|}
\hline
$K$     &                    & Cluster  & Rouge 1              & Rouge 2              & Rouge L              & Rouge L Sum          \\
        &                    & Mapping  & & & & \\
\hline
                             {2} &    & closest & 38.47(0.08)          & 16.55(0.06)          & 26.49(0.06)         & 34.17(2.41)          \\
                  &                   & least   & 37.83(1.65)          & 16.34(0.60)          & 26.05(1.11)          & 33.72(3.73) \\ 
                  &                   & longest   & \textbf{39.22(0.07)}          & \textbf{16.92(0.06)}          & \textbf{26.99(0.06)}          & \textbf{34.83(2.43)} \\ 
                  &                   & shortest   & 37.94(0.04)          & 16.24(0.04)          & 26.18(0.04)          & 33.70(2.39) \\ \cline{1-7}
                   
                             {3} &    & closest & 30.24(0.16)          & 13.02(0.16)          & 20.80(0.15)          & 28.26(0.15)          \\
                  &                   & least   & 30.30(0.34)          & 13.09(0.22)          & 20.84(0.26)          & 28.30(0.31) \\ 
                  &                   & longest   & \textbf{30.76(0.08)}          & \textbf{13.30(0.13)}          & \textbf{21.16(0.10)}          & \textbf{28.74(0.07)} \\ 
                  &                   & shortest   & 30.09(0.07)          & 12.96(0.12)          & 20.74(0.09)          & 28.12(0.07) \\ \cline{1-7}

                             {4} &    & closest & 26.42(0.12)          & 11.28(0.04)          & 18.19(0.08)          & \textbf{24.69(0.12)}          \\
                  &                   & least   & 26.68(0.08)          & 11.44(0.03)          & 18.38(0.05)          & 23.76(1.62) \\ 
                  &                   & longest   & \textbf{26.86(0.10)}          & \textbf{11.52(0.04)}          & \textbf{18.49(0.06)}          & 23.91(1.62) \\ 
                  &                   & shortest   & 26.34(0.11)          & 11.25(0.05)          & 18.16(0.07)          & 23.46(1.60) \\ \cline{1-7}
                  
                  \hline
\end{tabular}
        \caption{The performance of baseline on different cluster mapping methods with different summary numbers and random sampling method. The std is reported in the bracket.}
        \label{tab:two-step cluster reduction}
    \end{table*}



\begin{table}[!ht]
\centering
\begin{tabular}{c|llllll}
\hline
    & Article                    & Cluster                   & Rouge 1              & Rouge 2              & Rouge L              & Rouge L Sum          \\ 
    & Selection                  & Mapping & & & & \\ \hline

\multirow{3}{*}{} & \multirow{3}{*}{random}   & shortest & 38.50(0.11)          & 16.38(0.10)          & 26.64(0.13)          & 35.77(0.10)          \\
                  &                           & longest  & 38.51(0.12)          & 16.38(0.10)          & 26.64(0.13)          & 35.77(0.10)          \\
                  &                           & closest  & 38.51(0.11)          & 16.38(0.10)          & 26.64(0.13)          & 35.77(0.10)          \\ \cline{2-7}
                  & \multirow{3}{*}{selected} & shortest & 38.49(0.23)          & 16.30(0.20)          & 26.77(0.20)          & 30.94(2.89)          \\
                  &                           & longest  & 38.48(0.21)          & 16.30(0.20)          & 26.76(0.19)          & 30.92(2.87)          \\
                  &                           & closest  & 38.50(0.23)          & 16.30(0.20)          & 26.77(0.20)          & 30.95(2.89)          \\ \hline
\end{tabular}
\caption{The performance of JADS on different article selection methods for $K=2$. The std is reported in the bracket.}
\label{tab:sampling_method1}
\end{table}

    \begin{table*}[!ht]
        \centering
\begin{tabular}{c|llllll}
\hline
$K$ &                     & Cluster                   & Rouge 1              & Rouge 2              & Rouge L              & Rouge L Sum          \\ 
    &                       & Mapping & & & & \\ \hline

\multirow{3}{*}{2}   &    & shortest & 38.50(0.11)          & 16.38(0.10)          & 26.64(0.13)          & 35.77(0.10)          \\
                     &                           & longest  & 38.51(0.12)          & 16.38(0.10)          & 26.64(0.13)          & 35.77(0.10)          \\
                     &                           & closest  & 38.51(0.11)          & 16.38(0.10)          & 26.64(0.13)          & 35.77(0.10)          \\\cline{2-7}
                     
\multirow{3}{*}{3}   &    & shortest & 37.33(0.09)          & 15.61(0.11)          & 25.94(0.02)          & 34.68(0.06)          \\
                     &                           & longest  & 37.33(0.09)          & 15.61(0.11)          & 25.94(0.02)          & 34.68(0.07)          \\
                     &                           & closest  & 37.33(0.09)          & 15.61(0.11)          & 25.94(0.02)          & 34.68(0.07)          \\\cline{2-7}
                     
\multirow{3}{*}{4}   &    & shortest & \textbf{36.38(0.17)} & \textbf{14.98(0.15)} & 25.24(0.19)          & 33.77(0.17)          \\
                     &                           & longest  & 36.37(0.18)          & 14.97(0.15)          & 25.24(0.19)          & 33.77(0.17)          \\
                     &                           & closest  & \textbf{36.38(0.17)} & \textbf{14.98(0.15)} & 25.24(0.19)          & 33.77(0.17)          \\\cline{2-7}
                    
\multirow{3}{*}{Var} &    & shortest & 35.84(0.40)          & 14.93(0.14)          & 24.84(0.26)          & 33.27(0.37)          \\
                     &                           & longest  & 35.94(0.47)          & 14.99(0.18)          & 24.90(0.31)          & 33.36(0.43)          \\
                     &                           & closest  & \textbf{36.04(0.51)} & \textbf{15.05(0.20)} & \textbf{24.94(0.32)} & \textbf{33.46(0.46)} \\\cline{2-7}
                     
\end{tabular}
        \caption{The performance of JADS on different cluster mapping methods with different summary numbers and random sampling. The std is reported in the bracket.}
        \label{tab:model_reduction_method}
    \end{table*}

        \begin{table*}[!ht]
            \centering
\begin{tabular}{l|lll}
\hline
$K$          & Method      & \# of clusters        & F1                   \\ \hline
\multirow{3}{*}{2} & longest     & \textbf{2.00(0.00)}   & \textbf{0.88(0.15)}          \\
             & shortest    & 2.00(0.00)            & 0.85(0.19) \\ 
             & closest     & 1.99(0.09)            & 0.86(0.17) \\ 
             & least       & 1.94(0.14)            & 0.85(0.18) \\ \hline
\multirow{3}{*}{3} & longest     & \textbf{2.40(0.49)}   & \textbf{0.52(0.13)}          \\
             & shortest    & 2.40(0.49)            & 0.51(0.13) \\ 
             & closest     & 2.39(0.50)            & 0.51(0.13) \\ 
             & least       & 2.37(0.52)            & 0.52(0.13) \\ \hline
\multirow{3}{*}{4} & longest     & \textbf{2.85(0.81)}   & \textbf{0.57(0.18)}          \\
             & shortest.   & 2.85(0.81)            & 0.55(0.18) \\ 
             & closest     & 2.84(0.81)            &  0.56(0.18) \\ 
             & least       & 2.84(0.82)            & 0.56(0.19) \\ \hline
\end{tabular}
            \caption{The analysis of the clustering quality for baseline cluster mapping methods.}
            \label{tab:clustering_analysis_two_step_methods}
        \end{table*}

    \begin{table}[]
        \scalebox{0.7}{
        \centering
        \begin{tabular}{p{1cm}|p{5cm}|p{5cm}|p{5cm}|p{5cm}}
\hline
  $K$  & label          & baseline          & JADS               & pre-trained JADS         \\
\hline
  0 & Pit crew member Todd Phillips was
hit by a car on Sunday during the
inaugural IndyCar Grand Prix of
Louisiana. He was injured when he
was struck by the car of Francesco
Dracone, who had come in on Lap 25
for tires and fuel. Phillips
received stitches for a cut on his
leg and has been released. Dracone
did not finish the race and wound
up 23rd.                & Tim Sherwood's Aston Villa beat
Tottenham 1-0 at White Hart Lane.
Yannick Bolasie scored three as
Crystal Palace beat Sunderland 4-1.
Hull City remain in danger
following a 2-0 loss at
Southampton. Arsenal kept up their
impressive recent form with a 1-0
win over Burnley. Wilfried Zaha
scored a hat-trick for Crystal
Palace against Sunderland. Yannick
Bolasie became first Crystal Palace
player to score a hat-trick in
Premier League. Yannick Bolasie
became first Palace player to score
a hat-trick                   & Yannick Bolasie scored three as
Crystal Palace beat Sunderland 4-1.
Yannick Bolasie scored a hat-trick
as Hull City beat Southampton 2-0.
Christian Benteke scored a late
winner for Aston Villa against
Tottenham.                   & Yannick Bolasie scored a hat-trick
for Palace against Sunderland. Tim
Sherwood's side came from behind to
beat West Brom 2-0 at White Hart
Lane. Yannick Bolasie was the first
Palace player to score a hat-trick
in the Premier League.                    \\ \hline
  1 & Group released in northern Iraq
made up of 40 children, women and
elderly. Were piled onto a minibus
that then drove them to peshmerga
positions. Prisoners spent nearly a
year in Isis captivity, Kurdish
military has said.
No explanation has been given as to
why the 216 Yazidis were released.                & Yazidis, made up of women,
children and elderly, are said to
be in poor health and bearing signs
of abuse and neglect. They were
piled onto minibuses that drove
them to peshmerga positions. Some
of the prisoners are said to be in
poor health and bearing signs of
abuse and neglect. Yazidi community
leaders were there to receive the
freed women, children, elderly and
disabled. No reason was given for
the decision by ISIS.                   & The 216 prisoners have been held
captive in the Islamic State since
last summer. They were taken away
by ambulances and buses to receive
treatment. The freed prisoners said
they had been led to their
execution. But instead, they were
piled onto minibuses that drove
them to peshmerga positions. Yazidi
activists say many remain in the
hands of Islamic State.                   & The 216 prisoners have been held
captive in the Islamic State since
last summer when militants attacked
their villages in the area around
Sinjar in northwestern Iraq. The
freed Yazidis were taken away by
ambulances and buses to receive
treatment and care. Some of the
freed Yazidis are said to be in
poor health and bearing signs of
abuse and neglect. Yazidi activists
say many remain in the hands of
Islamic State, which has often
subjected women to rape or sexual
slavery.                    \\ \hline
  2 & Bolasie scored three goals as
Palace beat Sunderland 4-1. Villa
recorded a crucial 1-0 win at
Tottenham to stay above drop zone.
Aaron Ramsey scored Arsenal's goal
in 1-0 win at Burnley. Jamie Vardy
struck late to give Leicester a
vital 3-2 win at West Brom.                & Homes such as these in China are
known as 'dingzihu' or 'nail
houses' because they stick out and
are difficult to remove. One house
in Wenling, Zhejiang province, had
a main road built around it when
the owner refused to move. Another
image shows a house sitting alone
in a crater at the centre of a
construction site in Chongqing
Municipality.                   & Todd Phillips, 40, was struck by a
car as he flags down his driver
during Sunday's inaugural Indy
Grand Prix of Louisiana. The pit
crew worker was hit by the back end
of the car and was taken to the
infield care center for treatment
where he was swiftly given the all-
clear. Phillips only sustained
minor injuries to his leg which
required six stitches.                   & Surveillance footage shows
40-year-old Todd Phillips being
struck at full speed as he flags
down one of his drivers during
Sunday's inaugural Indy Grand Prix
of Louisiana. As the Dale Coyne
Racing team chief mechanic is hit
on the leg by the back end of the
vehicle he flips forwards and
performs a somersault before co-
workers rush over to check he's
okay. Amazingly Phillips of
Franklin, Wisconsin, only sustained
minor injuries to his leg which
required six stitches.                    \\ \hline
  3 & Homes such as these in China are
known as 'nail houses' because they
are difficult to remove, like a
stubborn nail. One house in
Wenling, Zhejiang province, had a
main road built around it when the
owner refused to move. Another
image shows a house sitting alone
in a crater at the centre of a
construction site in Chongqing
Municipality.                & Todd Phillips, 40, was struck by
Francesco Dracone, who had come in
on Lap 25 for tires and fuel.
Phillips was taken to the infield
care center for treatment where he
was swiftly given the all-clear.
Phillips said that in almost 20
years of being on the track he had
never been hit by a race car
before.                   & Homes such as these in China are
known as 'dingzihu' or 'nail
houses' because they stick out and
are difficult to remove. The owner
of the house didn't reach an
agreement with the local authority
about compensation of the
demolition.                   & The 17-race IndyCar season will
conclude on August 30 at the Sonoma
Raceway in California - one week
before the Labor Day Weekend
holiday. The 17-race IndyCar season
will conclude on August 30 at the
Sonoma Raceway in California.                    \\ \hline
  4 & mean\_rouge\_sum & 49.37 & 33.63 & 17.43 \\
\hline
\end{tabular}}
        \caption{Example where baseline has highest rouge mean score. }
        \label{tab:best_baseline}
    \end{table}
    
        \begin{table*}[!ht]
        \centering
\begin{tabular}{l|l|llll}
\hline
$K$                   &          Shuffling Methods              & Rouge 1              & Rouge 2              & Rouge L              & Rouge L Sum          \\ \hline
\multirow{3}{*}{2} & no-shuffling           & \textbf{40.56(0.19)} & \textbf{18.22(0.17)} & \textbf{28.64(0.12)} & 32.79(2.79)          \\
                   & in-sample-shuffling    & 38.96(0.20)          & 16.75(0.14)          & 27.09(0.10)          & \textbf{36.22(0.17)} \\
                   & cross-sample-shuffling & 38.51(0.11)          & 16.38(0.10)          & 26.64(0.13)          & 35.77(0.10)          \\ \hline
\multirow{3}{*}{3} & no-shuffling           & \textbf{40.96(0.19)} & \textbf{18.58(0.10)} & \textbf{28.89(0.19)} & 36.09(1.45)          \\
                   & in-sample-shuffling    & 38.88(0.18)          & 16.75(0.12)          & 27.10(0.15)          & \textbf{36.12(0.19)} \\
                   & cross-sample-shuffling & 37.33(0.09)          & 15.61(0.11)          & 25.94(0.02)          & 34.68(0.07)          \\ \hline
\multirow{3}{*}{4} & no-shuffling           & \textbf{40.81(0.05)} & \textbf{18.39(0.05)} & \textbf{28.67(0.06)} & 32.90(2.81)          \\
                   & in-sample-shuffling    & 38.68(0.51)          & 16.57(0.28)          & 26.96(0.28)          & \textbf{35.94(0.54)} \\
                   & cross-sample-shuffling & 36.38(0.17)          & 14.98(0.15)          & 25.24(0.19)          & 33.77(0.17)          \\ \hline
\end{tabular}
            \caption{The performance of JADS with different shuffling methods. The reported score is tested on ``random-choosing'' dataset, and the standard deviation is reported in the bracket}
            \label{tab:performance with different shuffling}
        \end{table*}

        \begin{table*}[!ht]
            \centering
\begin{tabular}{l|llll}
\hline
$K$            & 1           & 2           & 3           & 4           \\ \hline
Rouge 1     & \textbf{42.68} & 40.56 & 40.96 & 40.81 \\
Rouge 2     & \textbf{19.48} & 18.22 & 18.58 & 18.39 \\
Rouge L     & \textbf{29.61} & 28.64 & 28.89 & 28.67 \\
Rouge L Sum & \textbf{40.15} & 32.79 & 36.09 & 32.90 \\ \hline
\end{tabular}
            \caption{Performance of JADS with different summary number on ``random-choosing\_no-shuffling'' dataset. ``1'' means the model fine-tuned and tested on the original CNN/DM dataset}
            \label{tab:performance with summary number}
        \end{table*}

\clearpage

\begin{table}[!ht]
    \scalebox{0.7}{
    \centering
    \begin{tabular}{p{1cm}|p{5cm}|p{5cm}|p{5cm}|p{5cm}}

\hline
  $K$  & label          & baseline          & JADS              & pre-trained JADS        \\
\hline
 0 & Michael Vaughan is the hot
favourite to replace Paul Downton.
Vaughan  named on shortlist along
with Andrew Strauss and Alec
Stewart. The 2005 Ashes winning
captain has held talks with Tom
Harrison.                & Michelle Filkins, 44, of West
Wareham has been charged with
breaking and entering, larceny over
\$250, and malicious destruction of
property. She was discovered at the
Court Street property in Edgartown
by owner Mark Conklin on April 17.
When he confronted her she claimed
that she owned the house. She then
fled in a gray Nissan pickup truck
when Conklin called the police at
about 8 a.m. When police arrived,
they discovered a large glass
window to the left side of the
front door had been broken and
covered by a blanket. She is next
scheduled to be                    & Virgil van Dijk believes he sampled
the worst refereeing decision of
his career the night he landed an
early red card against Inter Milan.
The Dutchman believes the decision
was made 'terrible' Ronny Deila
described the defeat as the worst
of his career.                   & Virgil van Dijk believes he sampled
the worst refereeing decision of
his career the night he landed an
early red card against Inter Milan.
The Dutchman found sleep elusive
after a contentious Scottish Cup
semi-final defeat to Inverness
Caledonian Thistle. Ronny Deila
described the circumstances of the
defeat as the worst of his career.                   \\ \hline
  1 & Michelle Filkins, 44, of West
Wareham has been charged with
breaking and entering, larceny over
\$250, and the malicious destruction
of property. She was arrested on
April 17 after owner Mark Conklin
found her sitting in his summer
home. A neighbor told police he saw
Filkins outside with items from the
house and that she appeared to be
having a yard sale or giving the
items away. Police are asking
anyone who received items from the
home - including a lamp and a
painting - to return them.                & Michael Vaughan is the favourite
to become England cricket director.
The 2005 Ashes winning captain is
the hot favourite on the three-man
shortlist compiled by the ECB.
Andrew Strauss and Alec Stewart are
both registered their interest.
Peter Moores has said he could work
with Vaughan despite their uneasy
relationship.                    & Michael Vaughan has confirmed he
has held talks with the ECB over
becoming England's new cricket
director. Vaughan has worked
closely with incoming ECB chairman
Colin Graves. Andrew Strauss and
Alec Stewart are both in the frame
to take the role. Vaughan has
worked closely with Giles at
Yorkshire and is close to
appointing Giles.                   & Angelina Santini from San Diego,
California, filmed her nine-month-
old son Marcus getting carried away
in his bouncer one day. At the one-
minute-30-second mark, Marcus shows
no sign of slowing down. The
comical clip shows the youngster
lurching back and forth with the
device almost touching the floor.                   \\ \hline
  2 & Angelina Santini from San Diego,
California, filmed her son Marcus
getting carried away in his bouncer
one day. The comical clip shows the
youngster lurching back and forth
with the device almost touching the
floor.                & Celtic are eight points clear at
the top of the Scottish
Championship. Celtic have won the
last five games and are eight
points clear at the top. The Hoops
are currently playing in the
Premier League.                    & Angelina Santini from San Diego,
California, filmed her nine-month-
old son Marcus getting carried away
in his bouncer one day. The comical
clip shows the youngster lurching
back and forth with the device
almost touching the floor.                   & Michael Vaughan has confirmed that
he has held talks with the ECB over
becoming England’s new head of
cricket. The 2005 Ashes winning
captain is the hot favourite on the
three-man shortlist compiled by the
ECB at Thursday’s management board
meeting ahead of Andrew Strauss and
Alec Stewart, who have both
registered their interest. Vaughan
spoke to the ECB about the role on
Thursday after emerging as the No 1
choice.                   \\ \hline
  3 & Celtic crashed lost to 3-2 to
Inverness in the Scottish Cup semi-
final. The referee failed to spot a
handball from Inverness player Josh
Meekings. The decision denied
Celtic a penalty during a crucial
point of the game. Van Dijk was
sent off against Inter Milan in the
last 32 of the Champions League in
February.                & 'I was maybe near the halfway
line. Even in the stands you could
see it. It was a miracle that we
didn't get the right decision. It
was terrible. I don't know - I've
experienced some decisions lately.
The decision was terrible. The
Treble was very close. I go on
holiday so I don't think i will see
it,' he said. 'I'm focusing on the
present. His dilemma is clear. It
is what it is right now. It is what
it is right now. It is what it is
right now. It is what it is right
now. It is what it                    & Michelle Filkins, 44, of West
Wareham has been charged with
breaking and entering, larceny over
\$250, and malicious destruction of
property. Filkins was arrested on
April 17 after owner Mark Conklin
found her sitting in his summer
home. She was discovered at the
Court Street property in Edgartown
by owner Mark Conklin on April 17.                   & Michelle Filkins, 44, of West
Wareham has been charged with
breaking and entering, larceny over
\$250, and malicious destruction of
property. She was discovered at the
Court Street property in Edgartown
by owner Mark Conklin on April 17.
Filkins was arrested several hours
later after she attempted to get
back into the home thinking that
Conklin had left. Police are asking
anyone who received items from the
home - including a lamp and a
painting - to return them to the
Edgartown police department.                   \\ \hline
  4 & mean\_rouge\_sum & 28.81 & 50.96 & 51.53 \\
\hline
\end{tabular}}
    \caption{Example where JADS has highest rouge mean score.}
    \label{tab:JADS_best}
\end{table}

\clearpage

\begin{table}[!ht]
    \scalebox{0.7}{
    \centering
    \begin{tabular}{p{1cm}|p{5cm}|p{5cm}|p{5cm}|p{5cm}}

\hline
   $K$  & label          & baseline          & JADS               & pre-trained JADS        \\
\hline

  0 & Former NFL cornerback Will Allen
and his business partner Susan Daub
are facing civil fraud charges from
federal regulators. They allegedly
reaped over \$31 million in a Ponzi
scheme that promised high returns
to investors from funding loans to
cash-strapped pro athletes. Allen,
36, was a cornerback in the NFL
from 2001 to 2012, playing for the
New York Giants and the Miami
Dolphins. The SEC said Allen and
Daub paid about \$20 million to
investors but received only around
\$13 million in loan repayments from
athletes. To make up the gap they
paid investors with other
investors' money rather than actual
profits on the investments, the
agency said.                & Seven-day-old foal nuzzles Sunny
Bayne's shoulder before pushing her
to the ground and lying on top of
her belly. The young rider from
Kentucky can't stop smiling at the
animal's silly antics. To date the
clip of Bayne has received
thousands of hits online. To date
the clip of Bayne has received
thousands of hits online.                   & Will Allen and Susan Daub are
accused of running a multi-million
dollar Ponzi scheme. The SEC
alleges that Allen and Daub used
some funds from investors to cover
personal charges and finance other
business ventures. The pair are not
represented in the case by
attorneys, according to the SEC.                    & Photographers captured some
remarkable pictures of the Lyrid
meteor shower above the UK last
night. The meteor shower, visible
around the world but best seen from
Europe, peaked overnight with
between ten and 20 an hour. The
strength of the showers vary from
year to year and most years there
are no more than five to 20 meteors
an hour.                   \\ \hline
  1 & Footage shows the seven-day-old
foal nuzzling Sunny Bayne's
shoulder before pushing her to the
ground and lying on top of her
belly. The young rider from
Kentucky can't stop smiling at the
animal's antics. To date the clip
of Bayne has received thousands of
hits online.                & William D. Allen, 36, was a
cornerback in the NFL from 2001 to
2012, playing for the New York
Giants and the Miami Dolphins. He
was signed by the New England
Patriots in March 2012 but was
placed on injured reserve the
following August, and he left
football in March 2013. The SEC
said Allen and Daub paid about \$20
million to investors but received
only around \$13 million in loan
repayments from athletes. They used
some funds from investors to cover
personal charges at casinos and
nightclubs and to finance other
business ventures, the agency
alleged.                   & Karlis Bardelis, 30, was
attempting to scale the cliff in
Coire an Lochain, Scotland, when
the accident happened. Footage
shows the experienced climber
tumbling down the mountain head-
first. The experienced Latvian
climber, who now works in Nepal,
held his nerve after the tumble and
even continued his ascent to the
top.                    & Karlis Bardelis, 30, was
attempting to scale the cliff in
Coire an Lochain, Scotland, when
the accident happened. He was
rescued by his safety rope after
his axe became dislodged from the
ice, knocking the other out of
position and sending him plunging
down the mountain.                   \\ \hline
  2 & Photographers in the UK captured
the Lyrid meteor shower in the sky
last night. It occurs every year
around 16 to 25 April, so you can
still catch some meteors tonight
and tomorrow. The strength of the
showers vary from year to year and
most years there are no more than
20 meteors an hour. But in 1982
Americans counted nearly 100 an
hour and in 1803 it was as high as
700 an hour.                &                   & Footage shows the seven-day-old
foal nuzzling Sunny Bayne's
shoulder before pushing her to the
ground and lying on top of her
belly. Bayne later identified the
owner of the foal as Florida-based
Meg Miranda. The video has received
thousands of hits online.                    & Footage shows the seven-day-old
foal nuzzling Sunny Bayne's
shoulder before pushing her to the
ground and lying on top of her
belly. She later identified the
owner of the foal as Florida-based
Meg Miranda. To date the clip of
Bayne has received thousands of
hits online.                   \\ \hline
  3 & Karlis Bardelis, 30, was scaling
the cliff face in Coire an Lochain,
Scotland. Camera shows his axe come
loose sending him tumbling down the
cliff. He's eventually rescued by
safety rope after a falling 10m in
3.5 seconds. Latvian Bardelis
remarkably escaped injury and
continued with his ascent.                &                   & Photographers captured some
remarkable pictures of the Lyrid
meteor shower above the UK last
night. The meteor shower has been
observed for the past 2,700 years
and peaked overnight with between
ten and 20 an hour. The Lyrids
occur when Earth passes through
debris left by Comet Thatcher,
which orbits the solar system every
415 years.                    & Former NFL cornerback Will Allen
and his business partner are facing
civil fraud charges from federal
regulators over allegations the
pair ran a multi-million dollar
Ponzi scheme. They're accused of
reaping more than \$31 million in a
Ponzi scheme that promised high
returns from funding loans to cash-
strapped pro athletes. The SEC said
Allen and his business partner
Susan Daub misled investors about
the terms and the existence of some
of the loans.                   \\ \hline
  4 & mean\_rouge\_sum & 34.47 & 40.98 & 56.33 \\
\hline
\end{tabular}}
    \caption{Example where pre-trained JADS has highest rouge mean score. }
    \label{tab:pre-trained_best}
\end{table}

\clearpage

\begin{table}[!ht]
    \scalebox{0.7}{
    \centering
    \begin{tabular}{p{1cm}|p{5cm}|p{5cm}|p{5cm}|p{5cm}}
\hline
 $K$   & label          & baseline           & JADS               & pre-trained JADS        \\
\hline

 0 & Injuries sustained in rugby can
range from bruises to spinal cord
damage. Academics say government
plans to increase school rugby
games is risky. Professor's son
suffered horrendous injuries
playing the sport aged 14. Allyson
Pollock wants to see an end to
tackling and scrums in the game.                & Children should not be encouraged
to play rugby at school, say
researchers. They say 1 in 8 will
need to miss at least seven
training sessions or matches.
Government wants rugby to be played
more at school as part of plans to
increase competitive sports, to
curb rising levels of obesity. But
Professor Allyson Pollock, an
expert in Public Health at Queen
Mary, University of London said
plans were 'extremely worrying'                   & Darren Bent scored in stoppage time
to rescue a 1-1 draw for Derby.
Alex Pritchard put Derby ahead in
the second minute of stoppage time.
Darren Bent equalised for the Rams
in the 92nd minute with a header.
Darren Bent scored the winner in
the 90th minute to rescue a point.                    & Darren Bent scored in stoppage time
to rescue a 1-1 draw for Derby. On-
loan Hull winger was caught by
Toumani Diagouraga in the fifth
minute. The on-loan Hull winger
curled the 25-yard free-kick over
the wall and just past David
Button's left post.                   \\\hline
  1 & Two neuroscientists have conducted
brain imaging to examine moments of
clarity. Sudden "insights" are
otherwise known as "Eureka" or
"Aha" moments. We can increase our
chance of these insights with a
variety of daily changes.                & Darren Bent scored in stoppage
time to rescue a draw for Derby.
Alex Pritchard put Brentford ahead
with a stunning strike. Tom Ince
twice went close in the opening
seven minutes. Derby were good
value for half-time lead but missed
chances.                   & Dr Lara Briden has been a
naturopathic doctor for 20 years.
She has been a naturopathic doctor
for 20 years. She says that PMS can
be banished with just a few simple
steps.                    & Dr Lara Briden, a naturopathic
doctor with nearly 20 years
experience in women's health,
recounts a patient who was
suffering with PMS whom she helped
by changing her eating habits. She
says that PMS can be banished
forever with just a few simple
steps.                   \\\hline
  2 & No-More-PMS diet consists of anti-
inflammatory foods and nutrients.
The eating plan was devised by
naturopathic doctor Lara Briden.
Research indicates that PMS is
caused by unhealthy hormone
receptors. Health of hormone
receptors is impaired by chronic
inflammation. Stress, smoking, and
eating certain foods are all causes
of inflammation. Cutting
inflammatory foods can result in
dramatic improvement in PMS.                & Unsweetened muesli with mixed
berries and unsweetened greek
yoghurt. Two scrambled eggs on
whole grain or gluten-free toast
with butter. Sp Spinach and goat
cheese frittata. Dr Briden's
suggested breakfasts in her 'No-
More-PMS' diet plan.                   & Dr Lara Briden has been a
naturopathic doctor for 20 years.
She says that PMS is caused by the
drop in hormones, which is caused
by the hormone receptor. This is
the hormone receptor that changes
hormone levels. This changes the
hormone receptor's ability to adapt
to changes in hormone receptors.
This can be a sign that your mind
is not in a good place, she says.                    & Research suggests that in trying
to conjure up inspiration, most of
us end up suppressing it.
Neuroscientists John Kounios and
Mark Beeman explain how to clear
out mental junk.                   \\\hline
  3 & Alex Pritchard puts Brentford
ahead with stunning curling shot.
Brentford dominate but fail to take
chances to add second goal. Darren
Bent pokes home from close range in
92nd minute.                & Pritchard scored the winner in the
21st minute. The goal was scored in
the second half. The goal was
scored in the second half.                   & New Zealand has a duty to protect
children from risks of injury, says
UN Convention on the Rights of the
Child. Under the UN Convention, the
governments have a duty to protect
children from risks of injury.                    & Green salads with oily fish and
sushi are some of the delicious and
nutritious lunch suggestions.
Roasted lamb shanks with potato
mash and steamed kale.                   \\\hline
  4 & mean\_rouge\_sum & 16.72 & 16.48 & 10.34 \\
\hline
\end{tabular}}
    \caption{Example where pre-trained JADS has the lowest rouge mean score. }
    \label{tab:pre-trained_jads_bad}
\end{table}

\clearpage

\begin{table}[!ht]
    \scalebox{0.7}{
    \centering
    \begin{tabular}{p{1cm}|p{5cm}|p{5cm}|p{5cm}|p{5cm}}
\hline
 $K$   & label          & baseline           & JADS               & pre-trained JADS         \\
\hline

  0 & Some are child actors, others are
former  Hollywood heartthrobs. Can
be attributed to poor lifestyles,
career lows, or shattered
relationships. Renée Zellweger
caused controversy with her
dramatically altered face.                & Some stars have taken a quicker
turn for the worst than their
contemporaries. From child actors
to supermodels, FEMAIL rounds up
the celebrities that have aged
terribly. From child stars to
celebrities who have had botched
surgeries, it seems some stars are
on a constant mission to stay
young.                    & FEMAIL rounds up the celebrities
that have aged terribly. Renée
Zellweger, Madonna, Pamela Anderson
and Michelle Williams all have
greyed with time. FEMAIL rounds up
the celebrities who have aged
terribly.                    & Paul Scholes says Manchester United
have nothing to play for at the end
of the season. Scholes believes
Angel di Maria's red card against
Arsenal allowed Louis van Gaal to
turn to his fringe players. The
51-year-old has been with wife
Angelina Jolie since 2005, with
whom he has six children with.                    \\ \hline
  1 & Photographer Atif Saeed crept to
within ten feet of the hungry male
lion. Fearless Mr Saeed left the
safety of his car to sit on the
ground for the shot. As soon as the
camera captured the image, Mr Saeed
had to flee for his life. The lion
charged at the intrepid
photographer as he closed the car's
door.                & LINDSAY LOHAN. JOHNNY DEPP. KATE
MOSS. BRAD PITT. BRITNEY SPEARS. I
Did It Again.                    & Atif Saeed, 38, from Lahore,
Pakistan was in the city's safari
park when he spotted the lion.
Armed only with a camera, he
managed to capture a couple of
frames.                    & Model and actress Sulinh
Lafontaine has repeatedly said she
worked as a stunt driver on the
Fast \& Furious franchise.
Lafontaine has about a dozen and
half minor movie credits –
including a manicurist in What Just
Happened and a hotel guest in Will
Smith’s Hitch. She listed Furious 7
as her most recent project on IMDB,
describing herself as an
unaccredited 'race car driver' in
the film. The 29-year-old star also
had run-ins with the law and was
charged with a DUI and possession
of drugs in 2007.                    \\ \hline
  2 & Sulinh Lafontaine presented
herself as a daredevil stunt driver
who appeared in newly released
blockbuster. Said in interview on
CNN iReport at Furious 7 premiere
in Hollywood that she was the only
female stuntwoman driving cars. The
production, in fact, employed seven
or eight female stunt drivers, but
Lafontaine was not one of them. She
only appeared as extra on the set
in a crowd of 1,500.                &                    & Paul Scholes believes Manchester
United have nothing to play for.
The Dutchman says United need to be
consistent throughout the season.
United have won 25 points from
their last 10 games.                    & Atif Saeed, 38, from Lahore,
Pakistan was in the city's safari
park when he spotted the male lion
in the distance. Armed only with a
camera, Mr Saeed retreated to
safety. Unfortunately, the clicking
sound of the camera's shutter
alerted the lion to Mr Saeed's
presence.                    \\ \hline
  3 & Angel di Maria's red card forced
Louis van Gaal to look at other
players. Juan Mata, Marouane
Fellaini and Ashley Young have been
in good form in recent matches.
Paul Scholes says United are
playing well when they have nothing
to play for.                &                    & Model and actress Sulinh
Lafontaine has repeatedly said she
worked as a stunt driver on Fast \&
Furious. But this week she was
unmasked as a fraud and a liar.
Lafontaine has about a dozen and
half minor movie credits –
including a manicurist in What Just
Happened and a hotel guest in Will
Smith’s Hitch. She listed Furious 7
as her most recent project on IMDB,
describing herself as an
unaccredited 'race car driver' in
the film.                    & FEMAIL rounds up the celebrities
that have aged terribly. From child
actors that swiftly deteriorated in
lime light or golden Hollywood
stars that have greyed with time,
FEMAIL rounds up the celebrities
that have aged terribly.                    \\ \hline
  4 & mean\_rouge\_sum & 2.53 & 18.80 & 25.09 \\
\hline

\end{tabular}}
    \caption{Example where baseline has lowest rouge mean score.}
    \label{tab:baseline_worst}
\end{table}

\clearpage

\begin{table}[!ht]
    \scalebox{0.7}{
    \centering
    \begin{tabular}{p{1cm}|p{5cm}|p{5cm}|p{5cm}|p{5cm}}
\hline
 $K$   & label          & baseline           & JADS               & pre-trained JADS         \\
\hline

  0 & Frankie Ruttledge was always the
'fat bridesmaid' for her friends.
The 24-year-old has since lost six
stone after adopting a healthy
diet. Frankie, from Yorkshire, is
now planning her own wedding next
year.                & Manchester City host Manchester
United at Old Trafford on Sunday.
City fans pine for the days when
Martin Edwards ruled the roost. Not
many City supporters look back
fondly on the chaotic era of Peter
Swales. Louis van Gaal has leaned
on assistant boss Ryan Giggs for
tips. Brian Kidd is more Mancunian
than Happy Mondays dancer Bez
drinking a can of Boddingtons in
the rain.                   & Belle Gibson has been a cheerleader
for many years but has now admitted
she never had cancer. She has
dropped from a size 22 to a size
10/12. She has dropped from a size
22 to a size 10/12.                   & Belle Gibson, a wellness warrior,
has admitted that she never had
cancer and does not want
forgiveness. She founded the Whole
Pantry - a book and app which over
300,000 people have downloaded.
Wendy L. Patrick PhD says we fell
for Belle because 'we love
heartwarming stories—especially
comeback stories' Born to Bond -
There is a Bit of Belle in All of
Us.                    \\ \hline
  1 & Wendy L. Patrick PhD is the author
of Red Flags: How to Spot
Frenemies, Underminers, and Toxic
People. Dr Patrick has outlined the
five reasons the we fell for Belle
Gibson's story. Belle Gibson
released a book and app about how
she beat terminal cancer. This week
she finally admitted she never
actually had cancer. She explains
that there is a bit of Belle in all
of us as we are born to bond. 'We
were Belle´s cheerleaders as she
recounted her fight against cancer'                & Belle Gibson, from New York, has
been diagnosed with cancer. She has
admitted she made her story up, but
we still believe in her. We are not
fooled by the rumours, we believe
they are lying. Here are five ways
we can overcome the truthfulness
bias.                   & Frankie Ruttledge, 24, from
Strensall, Yorkshire, weighed 18
stone 7lbs. The size 28 was spurred
on to lose weight by an aunt who
commented that she was'shocked' at
how big she had become. Frankie
joined Slimming World and started
losing weight. Dropped from 18st
7lbs to 12 stone in less than two
years.                   & Manchester City travel to Old
Trafford to face United on Sunday
afternoon. Louis van Gaal has tried
to immerse himself in the local
culture after taking over as
manager last summer. Manchester
City fans are derided by their
rivals for the amount of 'empty
seats' at the Etihad Stadium.                    \\ \hline
  2 & Manchester United host bitter
rivals City on Sunday afternoon. It
could be a derby without a single
local player on either side. Louis
van Gaal and Manuel Pellegrini have
both tried to immerse themselves in
the local culture. READ: Manchester
United end 499-day wait over
Manchester City. Vincent Kompany:
Beating United at Old Trafford will
help'rectify' season.                & Frankie Ruttledge, 24, weighed 18
stone 7lbs and struggled with her
weight. She ditched her diet of
crisps for breakfast, pork pies for
lunch and frozen pizza for dinner
and dropped to 12 stone in less
than two years. Now the Slimming
World consultant is just one year
away from walking down the aisle
for a fourth time.                   & Manchester City host Manchester
United in the derby on Sunday. The
derby is now one of the biggest
match-ups in world football. Louis
van Gaal has leaned on assistant
Ryan Giggs for tips. Manchester
City fans all from Stockport? No
Man City fans all from Manchester?
No Man City fans from Manchester?
No Man City fans from Manchester?
No Man City fans from Manchester?
Man City fans from Stockport                   & Frankie Ruttledge, 24, weighed 18
stone 7lbs and had struggled with
her weight all her life before
dieting to a svelte size 10/12.
Ditched her diet of crisps for
breakfast, pork pies for lunch and
frozen pizza for dinner and dropped
to 12 stone in less than two years.
She has dropped from a size 22 to a
size 10/12.                    \\ \hline
  3 & Iranian plane came within 50 yards
of U.S. Navy Sea Hawk copter. Navy
copter was on patrol in
international airspace. U.S.
official think Iranian plane may
have been under orders of local
commander.                & Iranian military observation
aircraft flew within 50 yards of an
armed U.S. Navy helicopter over the
Persian Gulf. The incident has
sparked concern that top Iranian
commanders might not be in full
control of local forces. The
incident took place as the U.S. and
other world powers meet with Iran
in Switzerland to negotiate a deal
limiting Tehran's nuclear program.                   &                   & The incident sparks concern that
top Iranian commanders might not be
in full control of local forces.
The Navy crew took photos of the
incident but the military is not
releasing them.                    \\ \hline
  4 & mean\_rouge\_sum & 21.02 & 10.32 & 25.84 \\
\hline

\end{tabular}}
    \caption{Example where JADS has lowest rouge mean score.}
    \label{tab:jads_worst}
\end{table}

\clearpage




\end{document}